\documentclass[runningheads]{llncs}

\usepackage[mobile]{eccv}

\usepackage{eccvabbrv}

\usepackage{graphicx}
\usepackage{booktabs}

\usepackage[accsupp]{axessibility}  

\usepackage{hyperref}

\usepackage{orcidlink}

\usepackage{multirow}
\usepackage{amsmath}
\usepackage{dsfont}
\usepackage{pifont}

\begin{document}

\title{Decomposition of Neural Discrete Representations for Large-Scale 3D Mapping} 

\titlerunning{DNMap}

\author{Minseong Park\orcidlink{0000-1111-2222-3333} \and
Suhan Woo\orcidlink{0009-0008-1947-750X} \and
Euntai Kim\thanks{Corresponding author}\orcidlink{0000-0002-0975-8390}}

\authorrunning{M.~Park et al.}

\institute{School of Electrical and Electronic Engineering, Yonsei University, Seoul, Korea\\
\email{\{msp922,wsh112,etkim\}@yonsei.ac.kr}}

\maketitle

\begin{abstract}
Learning efficient representations of local features is a key challenge in feature volume-based 3D neural mapping, especially in large-scale environments. In this paper, we introduce Decomposition-based Neural Mapping (DNMap), a storage-efficient large-scale 3D mapping method that employs a discrete representation based on a decomposition strategy. This decomposition strategy aims to efficiently capture repetitive and representative patterns of shapes by decomposing each discrete embedding into component vectors that are shared across the embedding space. Our DNMap optimizes a set of component vectors, rather than entire discrete embeddings, and learns composition rather than indexing the discrete embeddings. Furthermore, to complement the mapping quality, we additionally learn low-resolution continuous embeddings that require tiny storage space. By combining these representations with a shallow neural network and an efficient octree-based feature volume, our DNMap successfully approximates signed distance functions and compresses the feature volume while preserving mapping quality. Our source code is available at \url{https://github.com/minseong-p/dnmap}.
\end{abstract}

\section{Introduction}
\label{sec:intro}
\begin{figure}[t]
\centering
\includegraphics[width=.9\linewidth]{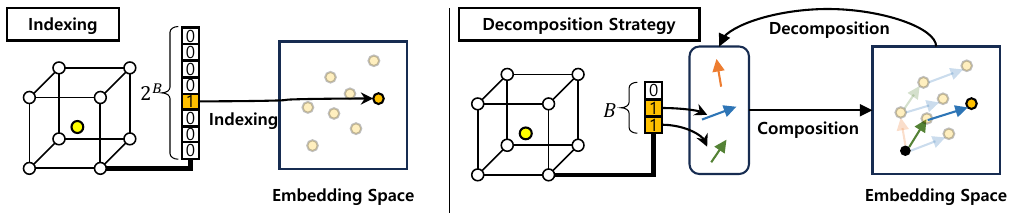}
\caption{The comparison between indexing~\cite{10.1145/3528233.3530727} and our proposed decomposition strategy. Indexing requires exploring the entire embedding space ($2^B$) to optimize discrete representations. In contrast, our approach only needs to learn $B$ indicators.}
\label{fig:decomp}
\end{figure}

Accurate and dense mapping in large-scale 3D environments plays a crucial role in the fields of computer vision and robotics, with a wide range of applications ranging from autonomous navigation to virtual environments. One of the most commonly used sensors for large-scale 3D mapping is LiDAR, known for its ability to provide accurate geometry and robustness to illumination changes. Due to these advantages, LiDAR is widely used for mapping in large-scale outdoor environments \cite{9197388, 6162880, 9981318}. A key practical challenge is balancing geometric accuracy with the memory consumption of the map. Since vehicles and mobile robots often have limited resources, including memory and processing power, efficient representation of maps is crucial to ensure that the map remains usable in these resource-constrained scenarios.

In recent years, implicit neural representations in 3D reconstruction have gained significant attention. Implicit neural representations learn a neural network to approximate implicit functions, such as signed distance functions (SDFs), enabling accurate, high-fidelity reconstruction using compact representations \cite{Park_2019_CVPR, Azinovic_2022_CVPR, Peng2020ECCV, Takikawa_2021_CVPR, mueller2022instant}. Due to these advantages, there have been several attempts to leverage implicit neural representations for 3D mapping using \cite{Yan_2021_ICCV, Azinovic_2022_CVPR, Zhu_2022_CVPR, Johari_2023_CVPR}. However, most of these methods only work in relatively small-scale indoor environments due to the limited capacity of neural networks or computational costs.

When it comes to large-scale outdoor environments, a few works have been reported. They have achieved some success in large-scale 3D mapping by learning and storing local features through efficiently designed feature volumes \cite{zhong2023icra, Deng_2023_ICCV, 10284988, 10238795}. However, they focused on designing the structure of the feature volume and relatively overlooked how to efficiently represent local features themselves. Understandably, as the scale of the environment increases, reducing the storage of each local feature becomes as critical as an efficient structure of the feature volume.

In this paper, we propose \textbf{D}ecomposition-based \textbf{N}eural \textbf{Map}ping (DNMap) to successfully adopt discrete representation in storage-efficient large-scale 3D mapping. Our DNMap approximates SDFs based on a decomposition-based discrete representation and octree-based structure. As shown in Fig.~\ref{fig:decomp}, we learn the decomposition and composition of the discrete embedding space, different from the existing discrete representation proposed in \cite{10.1145/3528233.3530727}. We decompose each discrete embedding into component vectors that are shared across the embedding space. Then, we learn how to compose the component vectors using composition indicators, which can be expressed as binary vectors. The process of composition is designed to have the same effect as indexing into multidimensional binary vectors, which are more computationally efficient than one-hot vectors, and the binary vectors are stored via an octree-based feature volume. Along with these techniques, our DNMap reconstructs accurate and compact large-scale outdoor environment maps by simultaneously optimizing the component vector set, the octree-based feature volume, and the decoder parameters.

Our main contributions are summarized as follows: 1) The proposed discrete representation based on a decomposition strategy significantly reduces the storage of the feature volume compared to continuous representation, while preserving mapping quality; 2) Learning composition, rather than indexing directly, reduces memory usage during training, enabling practical utilization of discrete representation in large-scale 3D mapping.

\section{Related Works}
\label{sec:related_works}

\subsection{Implicit Neural Representations}
Implicit neural representations approximate implicit surfaces as neural networks, enabling compact representation of complex environments. Early works on implicit neural representations represent an environment using a large, fixed single neural network \cite{Park_2019_CVPR, Azinovic_2022_CVPR, Peng2020ECCV, Takikawa_2021_CVPR, NEURIPS2021_e41e164f, nerf, zhang2020nerf}. Although there is an advantage that the amount of computation is constant regardless of the scale of the environment, the large amount of computations for inference was pointed out as a limitation. 

Recent works are based on the combination of a feature volume with a shallow neural network for scalable representations or real-time rendering \cite{Yan_2021_ICCV, Azinovic_2022_CVPR, Zhu_2022_CVPR, Johari_2023_CVPR, NEURIPS2020_b4b75896}. Most of these existing works have focused on representing objects or relatively small-scale indoor environments. There are several studies \cite{9811992, 10246371, 9477025, li2021semisupervised} on implicit representation in large-scale outdoor environments using LiDAR, but they focus on representation in single frames.

Implicit neural mapping, which aims to reconstruct 3D or visual maps using implicit neural representations is a relatively new task. Incremental mapping using a single neural network ~\cite{Yan_2021_ICCV, Yan_2023_ICCV} aims at representing sequential data in a replay-based manner. For large-scale 3D mapping using implicit neural representations, a few works \cite{zhong2023icra, Deng_2023_ICCV, 10284988, 10238795} focus on designing efficient data structures and have shown some success. However, we still think there are ways to represent features more efficiently, and this inspires our work.

\subsection{Neural Discrete Representation}
VQ-VAE~\cite{NIPS2017_7a98af17} was reported as a pioneering work to learn neural discrete representations. In this framework, the encoder outputs discrete latent codes via vector quantization (VQ), encouraging the network to generate meaningful latent codes, thereby enhancing the fidelity of reconstruction. Motivated by the VQ framework, VQ-AD~\cite{10.1145/3528233.3530727} was reported to compress feature volumes for an encoder-less framework, which is widely adopted in implicit representations. Despite its success in feature volume compression, VQ-AD raises memory usage issues during training because indexing discrete representations requires exploring the entire embedding space. This challenge greatly motivates our work.

\section{Preliminaries}
\label{sec:method}
In this paper, we follow the large-scale 3D mapping framework proposed in SHINE-Mapping~\cite{zhong2023icra}. Given a 3D environment $\mathcal{W}\subset{\mathbb{R}^3}$ sequentially observed by LiDAR in known poses, SHINE-Mapping approximates signed distance functions $d=\mathit{SDF}\left(\mathbf{x}\right)$ that map a 3D coordinate $\mathbf{x}\in\mathcal{W}$ to the shortest signed distance $d\in\mathbb{R}$ from the input coordinate to the surface.

Specifically, it consists of the following parts: an octree $\mathcal{O}$ with height $L$ is built from a given 3D environment $\mathcal{W}$; learnable embeddings are stored in voxel corners of an octree-based structure; given a sampled query point $\mathbf{x}$, the voxels to which the query point belongs are determined at each level, and the query feature $\mathbf{z}\in\mathbb{R}^D$ is determined as the sum of the embeddings interpolated from the voxel corners of each level, where $D$ indicates the dimension of the embedding space; the query feature $\mathbf{z}$ is decoded into a signed distance value through a decoder $f_{\Theta}$ which is a shallow MLP parameterized by $\Theta$. Then, the entire process $\Phi:\mathcal{W}\to\mathbb{R}$ can be represented as:
\begin{equation}
\label{eq:query_feature}
\mathbf{z}=\sum\limits_{l=0}^{L-1}{\mathit{interp}\left(\mathbf{x};\mathcal{O}_l\right)},
\end{equation}
\begin{equation}
\label{eq:func}
d=\Phi\left(\mathbf{x}\right)=f_\Theta\left(\mathbf{z}\right),
\end{equation}
where $\mathcal{O}_l$ is the set of embeddings of octree level $l$; $\mathit{interp}$ is a trilinear interpolation operator. Next, we will describe the details of the octree-based feature volume and the loss functions.

\subsubsection{Octree-Based Feature Volume.}
For a determined octree height $L$, SHINE-Mapping build an octree $\mathcal{O}$ from observed surfaces, allocating only the voxels that contain surfaces. The octree at each level consists of sparse voxels with sizes corresponding to each level. Based on the voxel size of the highest level, the lower the level, the larger the voxel size. The locations of voxels are stored as Morton codes. To speed up the mapping process, a hash table from voxels to the eight surrounding corners is maintained during mapping, and after mapping is complete, the hash table is removed to free storage space. After the octree is built, learnable embedding vectors are stored in the voxel corners.

\subsubsection{Loss Functions.}
The loss function $\mathcal{L}$ consists of the SDF loss function $\mathcal{L}_{\mathit{sdf}}$ and the Eikonal loss function $\mathcal{L}_{E}$. First, a binary cross entropy loss with a sigmoid function is applied as the SDF loss function $\mathcal{L}_{\mathit{sdf}}$ to supervise SDF prediction:
\begin{equation}
\label{eq:sdf_loss}
\mathcal{L}_{\mathit{sdf}}\left(\Phi\left(\mathbf{x}\right),\mathit{SDF}\left(\mathbf{x}\right)\right)= {\mathit{BCE}\left(\sigma\left(\Phi\left(\mathbf{x}\right)\right),\sigma\left(\frac{\mathit{SDF}\left(\mathbf{x}\right)}{s}\right)\right)},
\end{equation}
where $\mathit{BCE}\left(x,y\right)=y\log{x}+\left(1-y\right)\log{\left(1-x\right)}$ is the binary cross entropy loss, and $s$ is a hyper-parameter to control the flatness of the sigmoid function $\sigma\left(x\right)=1/\left(1+e^{-x}\right)$. This loss function emphasizes the impact of the query points sampled near the surfaces. Second, the Eikonal loss function $\mathcal{L}_e$ is adopted to encourage the L2-norm of the gradient at the query point $\mathbf{x}$ to equal one \cite{pmlr-v119-gropp20a}, since the gradient of SDF is a unit normal vector. The Eikonal loss function $\mathcal{L}_e$ is defined as:
\begin{equation}
\mathcal{L}_{e}\left(\Phi\left(\mathbf{x}\right)\right)=\left({\lVert{\nabla\Phi\left(\mathbf{x}\right)}\rVert}_2-1\right)^2.
\end{equation}
Consequently, the final loss $\mathcal{L}$ is defined as:
\begin{equation}
\mathcal{L}\left(\Phi\left(\mathbf{x}\right),\mathit{SDF}\left(\mathbf{x}\right)\right) = \mathcal{L}_{\mathit{sdf}}\left(\Phi\left(\mathbf{x}\right),\mathit{SDF}\left(\mathbf{x}\right)\right) + \lambda\mathcal{L}_{e}\left(\Phi\left(\mathbf{x}\right)\right),
\end{equation}
where $\lambda=0.1$ is weight of the Eikonal loss function $\mathcal{L}_e$.

\section{Method}
\begin{figure}[t]
\centering
\includegraphics[width=1.\linewidth]{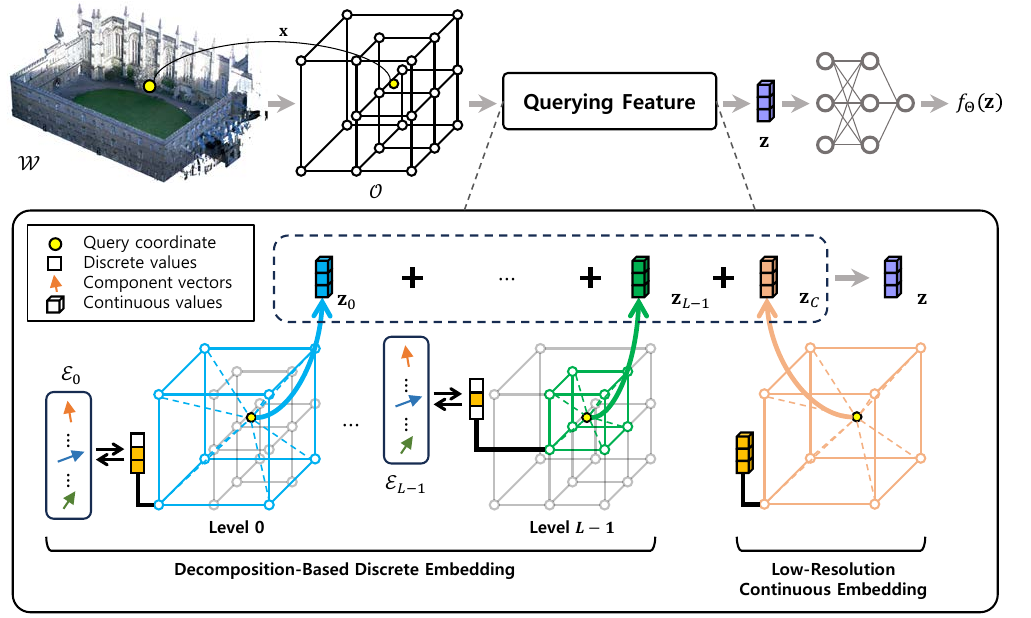}
\caption{The pipeline of our DNMap. When a query point $\mathbf{x}$ is given, the voxels to which the query point belongs are determined at each level of the octree $\mathcal{O}$. For the determined discrete embedding voxels at each level, the voxel corner embeddings are composed from the component vector set $\mathcal{E}$ according to the composition indicators stored in the corners. The query embeddings at each level $\mathbf{z}_0,\cdot\cdot\cdot,\mathbf{z}_{L-1}\mathbf{z}_c$ are determined via trilinear interpolation of the voxel corner embeddings. Finally, the query embedding $\mathbf{z}$ is determined as the sum of all interpolated features, and the decoder $f_\Theta$ takes the query embedding as input and outputs a signed distance value.}
\label{fig:overview}
\end{figure}

Advancing from SHINE-Mapping, we introduce \textbf{D}ecomposition-based \textbf{N}eural \textbf{Map}ping (DNMap) that learns discrete embeddings based on a decomposition strategy. Our DNMap differs from SHINE-Mapping in two key ways: composition indicators, rather than continuous embeddings, are stored in the voxel corners of the octree $\mathcal{O}$; a corner embedding becomes $\mathbf{e}_k$, composed from the component vector set $\mathcal{E}$. Additionally, we store low-resolution continuous embeddings in the voxel corners at the $0$-level of the octree to complement global location information. With these embeddings, the query feature $\mathbf{z}$ is determined as the sum of \textit{discrete} and \textit{continuous} terms, different from Eq.~(\ref{eq:query_feature}):
\begin{equation}
\label{eq:querying_feature}
\mathbf{z}=\underbrace{\sum\limits_{l=0}^{L-1}{\textit{interp}\left(\mathbf{x};\mathcal{D}_l\right)}}_{\text{discrete}}+\underbrace{\textit{interp}\left(\mathbf{x};\mathcal{C}\right)}_{\text{continuous}},    
\end{equation}
where $\mathcal{D}_l$ is a set of decomposition-based discrete embeddings which are composed from the component vector set $\mathcal{E}$ at level $l$; $\mathcal{C}$ is a set of low-resolution continuous embeddings. The framework of our DNMap is illustrated in Fig.~\ref{fig:overview}.

Finally, the objective of our DNMap is to simultaneously optimize the discrete feature volume  $\mathcal{D}$, low-resolution continuous feature volume $\mathcal{C}$, the component vector set $\mathcal{E}$, and the decoder parameter $\Theta$ by minimizing the final loss $\mathcal{L}$, and can be represented as:
\begin{equation}
\label{eq:objective}
\underset{\mathcal{D}, \mathcal{C},\mathcal{E},\Theta}{\mathop{\arg\min}}{\frac{1}{|\mathcal{W}|}\sum\limits_{\mathbf{x}\in\mathcal{W}}{\mathcal{L}\left(\Phi\left(\mathbf{x}\right),\mathit{SDF}\left(\mathbf{x}\right)\right)}}.
\end{equation}

\subsection{Decomposition-Based Neural Discrete Representation}
We define the discrete embedding space $\left\{\mathbf{e}_i\in\mathbb{R}^D\right\}_{i=0}^{2^B-1}$ at bitwidth $B$, where $\mathbf{e}_i$ is the $i$-th discrete embedding. A simple way to optimize the embedding space is to learn differentiable indices whose dimension is the size of the embedding space \cite{10.1145/3528233.3530727}. However, this approach has a problem when applied to large-scale 3D mapping because it requires extensive computation to explore the entire embedding space. To solve this problem, we adopt a simple decomposition strategy.

\subsubsection{Decomposition.}
We decompose the discrete embedding space into a set of $B+1$ component vectors, which consists of one embedding bias $\mathbf{e}_b\in\mathbb{R}^D$ and $B$ embedding offsets $\Delta{\mathbf{e}}\in\mathbb{R}^D$ that are shared across the embedding space. Then the component vector set $\mathcal{E}\subset{\mathbb{R}^D}$ is defined as: 
\begin{equation}
\label{eq:component_vector_set}
\mathcal{E}=\left\{\mathbf{e}_b,\Delta{\mathbf{e}}_0,\cdots,\Delta{\mathbf{e}}_{B-1}\right\}.
\end{equation}

\subsubsection{Composition.}
We represent a discrete embedding $\mathbf{e}$ as a composition of the component vectors, and the composition process is formulated as:
\begin{equation}
\label{eq:naive_composition}
\mathbf{e}=\mathbf{e}_b+\sum\limits_{j=0}^{B-1}{{b}_{j}\Delta{\mathbf{e}}_j},
\end{equation}
where $b_{j}\in\left\{0,1\right\}$ is the $j$-th composition indicator. 
To update the $j$-th embedding offset $\Delta{\mathbf{e}}_j$ by gradient descent, the gradient can be calculated using the chain rule:
\begin{equation}
\label{eq:basic_grad}
    \frac{\partial \mathcal{L}}{\partial \Delta{\mathbf{e}}_j} = \frac{\partial \mathcal{L}}{\partial d} \frac{\partial d}{\partial \mathbf{e}} \frac{\partial \mathbf{e}}{\partial \Delta{\mathbf{e}}_j}.
\end{equation}
Since the term $\frac{\partial \mathbf{e}}{\partial \Delta{\mathbf{e}}_j}$ in Eq.~(\ref{eq:basic_grad}) is not defined when $b_j=0$, we reformulate Eq.~(\ref{eq:naive_composition}) as follows so that each embedding offset is updated frequently:

\begin{equation}
\label{eq:composition}
\begin{split}
\mathbf{e}&=\underbrace{\mathbf{e}'_b+\sum\limits_{j=0}^{B-1}{\Delta{\mathbf{e}}^0_j}}_{\mathbf{e}_b}+\sum\limits_{j=0}^{B-1}{{b}_{j}\underbrace{\left(\Delta{\mathbf{e}}^1_j-\Delta{\mathbf{e}}^0_j\right)}_{\Delta{\mathbf{e}}_j}}\\
&=\mathbf{e}'_b+\sum\limits_{j=0}^{B-1}{\left\{\left(1-{b}_{j}\right)\Delta{\mathbf{e}}^0_j+{b}_{j}\Delta{\mathbf{e}}^1_j\right\}}.
\end{split}
\end{equation}

Then, the term $\frac{\partial \mathbf{e}}{\partial \Delta{\mathbf{e}}_j}$ can be defined as follows, regardless of $b_j$:
\begin{equation}
\label{eq:proposed_grad}
    \frac{\partial \mathbf{e}}{\partial \Delta{\mathbf{e}}_j} = 
    \begin{cases} 
        \displaystyle \frac{\partial \mathbf{e}}{\partial \Delta{\mathbf{e}}^1_j} \frac{\partial \Delta{\mathbf{e}}^1_j}{\partial \Delta{\mathbf{e}}_j} & \text{if } b_j = 1 \\
        \displaystyle \frac{\partial \mathbf{e}}{\partial \Delta{\mathbf{e}}^0_j} \frac{\partial \Delta{\mathbf{e}}^0_j}{\partial \Delta{\mathbf{e}}_j} & \text{if } b_j = 0
    \end{cases}
    .
\end{equation}
This approach helps the embedding space converge to an optimal representation of a large-scale 3D environment. 

\subsubsection{Composition indicator.}
We define a composition indicator vector that indicates which component vector to compose, which can be expressed as a binary vector $\mathbf{b}={\begin{bmatrix} b_0&\cdots&b_{B-1} \end{bmatrix}}^T\in\mathbb{R}^B$. Actually, the real gradient of a binary vector $\mathbf{b}$ is not defined because the binary vector is deterministic and non-differentiable. To approximate the gradient of the binary vector $\mathbf{b}$, we adopt a straight-through estimator approach~\cite{bengio2013estimating} using a sigmoid function $\sigma\left(\cdot\right)$ and an indicator function $\mathds{1}_{\mathbb{R}^{+}}\left(\cdot\right) $. The indicator function $\mathds{1}_{\mathbb{R}^{+}}:\mathbb{R}\to\left\{0,1\right\}$ is a function that outputs one if $x>0$, and zero otherwise. Specifically, we first store the randomly initialized differentiable continuous vectors $\mathbf{v}$ in the octree $\mathcal{O}$. When an index is called, we generate two types of vectors from the vector $\mathbf{v}$: non-differentiable binary vector $\mathds{1}_{\mathbb{R}^{+}}\left(\mathbf{v}\right)$ and differentiable vector $\sigma\left(\mathbf{v}\right)$. Then, the binary vector $\mathbf{b}$ is calculated as:
\begin{equation}
\mathbf{b} = \text{sg}\left[\mathds{1}_{\mathbb{R}^{+}}\left(\mathbf{v}\right) - \sigma\left(\mathbf{v}\right)\right] + \sigma\left(\mathbf{v}\right),
\end{equation}
where $\text{sg}\left[\cdot\right]$ the indicates stop-gradient operator, which is defined to treat a specific variable as a constant. This simple technique allows us to optimize the binary vector $\mathbf{b}$ via gradient descent.

\subsubsection{Implementation.} 
The composition process in Eq.~(\ref{eq:composition}) is implemented using a linear layer with $\mathbf{W}={\begin{bmatrix} \Delta{\mathbf{e}}^0_0 &\cdots&\Delta{\mathbf{e}}^0_{B-1} & \Delta{\mathbf{e}}^1_0 &\cdots&\Delta{\mathbf{e}}^1_{B-1} \end{bmatrix}}\in\mathbb{R}^{D\times{2B}}$ as the weight matrix and $\mathbf{e}'_b$ as the bias as follows:

\begin{equation}
\label{eq:implementation}
\mathbf{e}=\mathbf{W}\mathbf{b}^{*}+\mathbf{e}'_b
\end{equation}
where $\mathbf{b}^{*}\in\mathbb{R}^{2B}$ is the concatenation of two vectors, $1-\mathbf{b}$ and $\mathbf{b}$.

\subsection{Low-Resolution Continuous Embedding}
In order to complement mapping quality, we additionally learn low-resolution continuous embedding $\mathbf{c}\in\mathbb{R}^D$ which are stored in voxel corners at the 0-level octree to help the discrete embeddings that lack global location information. Because the number of voxels in the 0-level octree is a very small percentage of the number of voxels in all levels, it does not significantly affect the overall storage. For example, when building an octree with a height of 3 and a smallest voxel size of 0.2 m, the ratio of the number of 0-level voxels to the number of total level voxels in the MaiCity~\cite{vizzo2021icra} and NewerCollege~\cite{ramezani2020newer} datasets are 4.7\% and 3.7\%, respectively.

\subsection{Efficient Implementation}
\label{sec:efficient_implementation}
\begin{figure}[t]
\centering
\includegraphics[width=1.\linewidth]{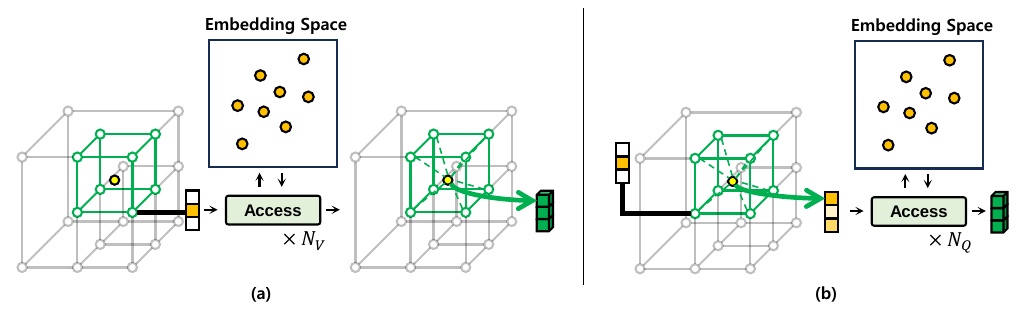}
\caption{Comparison between (a) basic implementation and (b) efficient implementation. In most cases during training, (b) is more efficient than (a) because $N_V>>N_Q$, where $N_V$ is the number of total voxels, and $N_Q$ is the number of query points processed at once.}
\label{fig:implementation_comparison}
\end{figure}
Because the region where the query points in the batch are located is part of a large-scale 3D environment, it is inefficient to compose all voxel corners every time during training. To address this problem, we first interpolate the binary vectors of the corners of the voxel to which the query point belongs and then perform indexing. Since the processes described in Eq.~(\ref{eq:querying_feature}) and Eq.~(\ref{eq:implementation}) are linear, the output query features are the same. This approach is more efficient because the number of query points in one batch is generally less than the total number of voxels. The $l$-level query feature with efficient implementation can be calculated as:
\begin{equation}
\label{eq:efficient_implementation}
\mathbf{z}_l=\mathbf{W}\textit{interp}\left(\mathbf{x};\mathcal{B}^*_l\right)+\mathbf{e}'_b,
\end{equation}
where $\mathcal{B}^*_l$ is the set of $\mathbf{b}^*$ at level $l$. This technique allows accessing the component vector set as many times as only the number of query points, which reduces redundant composition when querying features. Therefore, the execution time is reduced in most cases during training because the total number of voxels is much greater than the number of query points processed at once. A comparison between basic and efficient implementation is illustrated in Fig.~\ref{fig:implementation_comparison}.

\section{Experiments}

\subsection{Experimental Setting}
\label{sec:setting}
\subsubsection{Datasets.}
To validate our DNMap, we use two publicly available outdoor LiDAR datasets that provide ground truth mesh: MaiCity~\cite{vizzo2021icra}, and Newer College~\cite{ramezani2020newer}. MaiCity consists of 64-beam noise-free synthetic LiDAR scans in urban roads. Newer College consists of handheld LiDAR scans collected at Oxford University with cm-level noise and motion distortion.

\subsubsection{Data Preparation.}
To train SDFs, we prepare pairs of query points and SDF labels following SHINE-Mapping~\cite{zhong2023icra}. We sample query points along the ray of LiDAR point clouds and use the signed distance values from sampled query points to the endpoints of the rays as labels. Since our goal is to reconstruct surfaces, the query points sampled near the surface are more important than those sampled in free space. Therefore, we sample the query points near the surface and the query points in free space separately. Specifically, for each LiDAR ray, $N_F$ query points are uniformly sampled in free space, and $N_S$ query points inside the truncation band $\pm3s$ are sampled near the surfaces, where $s$ is a hyperparameter as described in Eq.~(\ref{eq:sdf_loss}).

\subsubsection{Evaluation Metric.}
To compare methods with each other, we measure metrics in terms of mapping quality. We use the commonly used 3D reconstruction metrics adopted in most works \cite{zhong2023icra}. We evaluate accuracy, completion, Chamfer-L1 distance, and F-score, obtained by comparing the reconstructed mesh with the ground truth mesh. The reconstructed mesh can be generated from learned implicit neural representations using marching cubes~\cite{10.1145/37402.37422}. We generate explicit mesh with a voxel size of 10 cm. Furthermore, we report the storage of neural representations and total storage, including the spatial model.

\subsubsection{Implementation Details.}
\label{sec:imp_detail}
For the spatial model, we set the height of the octree to $L=3$. For the representation, we set the embedding space dimension to $D=8$. For data preparation, we set $N_F=6$, $N_S=3$, and $s=0.05$ in MaiCity~\cite{vizzo2021icra} and $N_F=6$, $N_S=3$, and $s=0.1$ in Newer College~\cite{ramezani2020newer}. The decoder is implemented as a MLP containing two 32-dimensional hidden layers with ReLU activation. All methods are trained for 20000 iterations with 8192 batch size. An Adam optimizer is used with an initial learning rate 0.01, which decays to 0.001 after 10000 iterations.

\subsection{Experimental Results}
To verify the mapping quality and storage efficiency of our method, we compare our DNMap with SHINE-Mapping~\cite{zhong2023icra} and VQ-SHINE-Mapping~\cite{zhong2023icra,10.1145/3528233.3530727}. Based on the SHINE-Mapping framework, VQ-SHINE-Mapping~\cite{zhong2023icra,10.1145/3528233.3530727} is a method of replacing the original continuous embedding with the discrete embedding proposed in VQ-AD~\cite{10.1145/3528233.3530727}. We also report the results of DNMap-\textit{discrete}, which does not use low-resolution continuous embedding, to verify the effectiveness of the proposed discrete representation. All methods were evaluated in the same experimental setting and GPU, changing only the bitwidth from 4 to 8 when discrete embedding is used.

\subsubsection{Quantitative Comparisons.}
\begin{table}[t]
\caption{Quantitative comparisons on MaiCity~\cite{vizzo2021icra} dataset. The error threshold for calculating F-score is 10 cm.}
\label{tab:maicity}
\scriptsize
    \begin{center}
        \begin{tabular}{lccccccrrcrrc}
            \toprule
              & \quad & \multicolumn{4}{c}{Mapping Quality} & \quad & \multicolumn{2}{c}{Storage(kB)} & \quad & & &\quad\\ \cmidrule{2-10} 
             Method & & {Acc.(cm)$\downarrow$} & {Com.(cm)$\downarrow$} & {C-l1(cm)$\downarrow$} & {F-score$\uparrow$} & & Rep. & Total & & \multicolumn{2}{c}{Time} \\ \midrule
            \textbf{Discrete} (4-bit) \\
            VQ-SHINE-Mapping~\cite{zhong2023icra,10.1145/3528233.3530727} & & 4.57& 15.05& 9.81& 86.07& & 249 & 1937 & & 31m & 59s\\
            DNMap-\textit{discrete} & & \textbf{4.11}& \textbf{12.44}& \textbf{8.28}& \textbf{87.14}& & 248 & 1936 & & 30m & 04s\\ \midrule
            \textbf{Discrete} (6-bit) \\ 
            VQ-SHINE-Mapping~\cite{zhong2023icra,10.1145/3528233.3530727} & & 4.29& 15.47& 9.88& 87.33& & 376 & 2064 & & 34m & 40s\\ 
            DNMap-\textit{discrete} & & \textbf{4.12}& \textbf{12.21}& \textbf{8.16}& \textbf{87.64}& & 371 & 2059 & & 30m & 45s\\ \midrule
            \textbf{Discrete} (8-bit) \\ 
            VQ-SHINE-Mapping~\cite{zhong2023icra,10.1145/3528233.3530727} & & 4.19& 15.38& 9.79& \textbf{87.90}& & 519 & 2207 & & 43m & 40s\\ 
            DNMap-\textit{discrete} & & \textbf{4.08}& \textbf{12.42}& \textbf{8.25}& 87.63& & 496 & 2184 & & 30m & 57s\\ \midrule
            \textbf{Non-discrete} \\
            SHINE-Mapping~\cite{zhong2023icra} & & 3.19 & 17.75 & 10.47& 89.23& & 15775 & 17463 & & 30m & 03s\\ 
            DNMap (4-bit) & & 3.13& 17.52& 10.33& 89.45& & 1089 & 2777 & & 35m & 27s\\
            DNMap (6-bit) & & \textbf{3.11}& 16.96& 10.04& 89.55& & 1212& 2900 & & 35m & 46s\\
            DNMap (8-bit) & & 3.14& \textbf{16.46}& \textbf{9.80}& \textbf{89.56}& & 1337 & 3025 & & 36m & 00s\\
            \bottomrule
		\end{tabular}
	\end{center}
\end{table}

\begin{table}[ht]
\caption{Quantitative comparisons on Newer College~\cite{ramezani2020newer} dataset. The error threshold for calculating F-score is 20 cm. $^\star$ denotes out-of-memory.}
\label{tab:ncd}
\scriptsize
    \begin{center}
        \begin{tabular}{lccccccrrcrrc}
            \toprule
              & \quad & \multicolumn{4}{c}{Mapping Quality} & \quad & \multicolumn{2}{c}{Storage(kB)} & \quad & & &\quad\\ \cmidrule{2-10} 
             Method & & {Acc.(cm)$\downarrow$} & {Com.(cm)$\downarrow$} & {C-l1(cm)$\downarrow$} & {F-score$\uparrow$} & & Rep. & Total & & \multicolumn{2}{c}{Time} \\ \midrule
            \textbf{Discrete} (4-bit) \\
            VQ-SHINE-Mapping~\cite{zhong2023icra,10.1145/3528233.3530727} & & 8.57& \textbf{10.10}& 9.78& 91.31& & 662 & 6458& & 42m & 33s \\ 
            DNMap-\textit{discrete} & & \textbf{7.69}& 11.01& \textbf{9.35}& \textbf{92.86}& & 661 & 6457 & & 36m & 24s\\ \midrule
            \textbf{Discrete} (6-bit) \\
            VQ-SHINE-Mapping~\cite{zhong2023icra,10.1145/3528233.3530727} & & 8.00& \textbf{10.70}& 9.35& 92.01& & 995 & 6791& & 49m & 06s\\ 
            DNMap-\textit{discrete} & & \textbf{7.57}& 10.76& \textbf{9.16}& \textbf{92.83}& & 990 & 6786 & & 36m & 46s\\ \midrule
            \textbf{Discrete} (8-bit) \\
            VQ-SHINE-Mapping$^\star$~\cite{zhong2023icra,10.1145/3528233.3530727} & & -& -& -& -& & \multicolumn{1}{c}{-} & \multicolumn{1}{c}{-}& & \multicolumn{2}{c}{-}\\ 
            DNMap-\textit{discrete} & & \textbf{7.24}& \textbf{10.93}& \textbf{9.08}& \textbf{93.35}& &1321 &7117 & & 37m & 27s\\ \midrule
            \textbf{Non-discrete} \\
            SHINE-Mapping~\cite{zhong2023icra} & & 7.42& \textbf{10.47}& 8.94& 92.96& & 42187 & 47983 & & 35m & 58s\\ 
            DNMap (4-bit) & & 6.76& 10.50& 8.63& 93.63& & 2207 & 8003 & & 44m & 34s\\
            DNMap (6-bit) & & \textbf{6.58}& 10.50& \textbf{8.54}& \textbf{93.75}& & 2536 & 8332 & & 45m & 12s\\
            DNMap (8-bit) & & 6.68& 10.64& 8.67& 93.60& & 2867 & 8663 & & 45m & 27s\\
            \bottomrule
		\end{tabular}
	\end{center}
\end{table}

Tab.~\ref{tab:maicity} and Tab.~\ref{tab:ncd} show the 3D reconstruction results on the MaiCity~\cite{vizzo2021icra} dataset and Newer College~\cite{ramezani2020newer} dataset, respectively. We conducted the experiments by setting all methods to share the same spatial model with an octree with a height of 3 and a voxel size of 0.2 m. In the tables, \textbf{Discrete} indicates methods that use only discrete representations, and \textbf{Non-discrete} indicates methods that use continuous representations. In terms of storage, Rep. refers to the capacity used to store the feature, and the Total storage capacity includes Morton codes to represent the spatial model.

In the experiments on the MaiCity~\cite{vizzo2021icra} dataset, our DNMap outperforms all other methods even with less storage. Compared to VQ-SHINE-Mapping when the bitwidth is 8, our DNMap-\textit{discrete} shows slightly lower mapping quality in F-score, but trains about 1.4$\times$ faster. When comparing our DNMap (8-bit) with SHINE-Mapping, DNMap reconstructs a dense and accurate mesh even with 8.5\% of the feature storage.

In the experiments on the Newer College~\cite{ramezani2020newer} dataset, our DNMap outperforms SHINE-Mapping in all metrics, even in an environment with significant noise and motion distortion. These results imply that our DNMap represents large-scale environments with a limited number of component vectors, making it robust to noise. Also, when the bitwidth is 8, VQ-SHINE-Mapping cannot be trained due to large memory usage, whereas our DNMap can be trained. The weakness in the Newer College dataset is that our DNMap shows slightly lower performance in terms of completeness compared to other methods. Nevertheless, DNMap achieves better performance than other methods in all other metrics, especially when comparing our DNMap-\textit{discrete} (8-bit) with SHINE-Mapping. Even without using continuous embeddings, our DNMap reconstructs accurate mesh with only 3\% of the feature storage.

\subsubsection{Qualitative Comparisons.}
\label{sec:qualitative}

\begin{figure*}[hbt!]
\centering
  \begin{subfigure}{\linewidth}
    \centering
    \includegraphics[width=.9\linewidth]{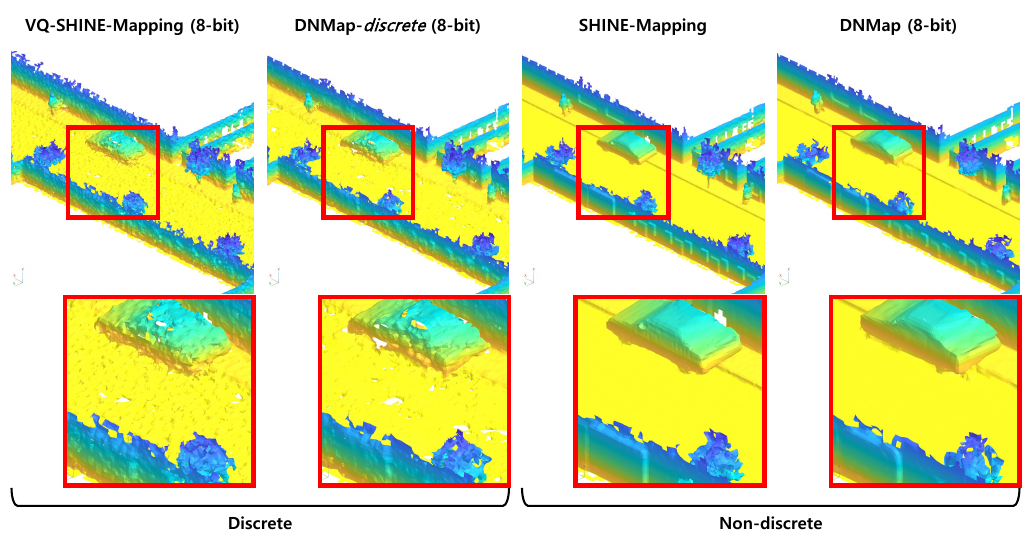}
    \caption{MaiCity}
  \end{subfigure}
  \begin{subfigure}{\linewidth}
    \centering
    \includegraphics[width=.9\linewidth]{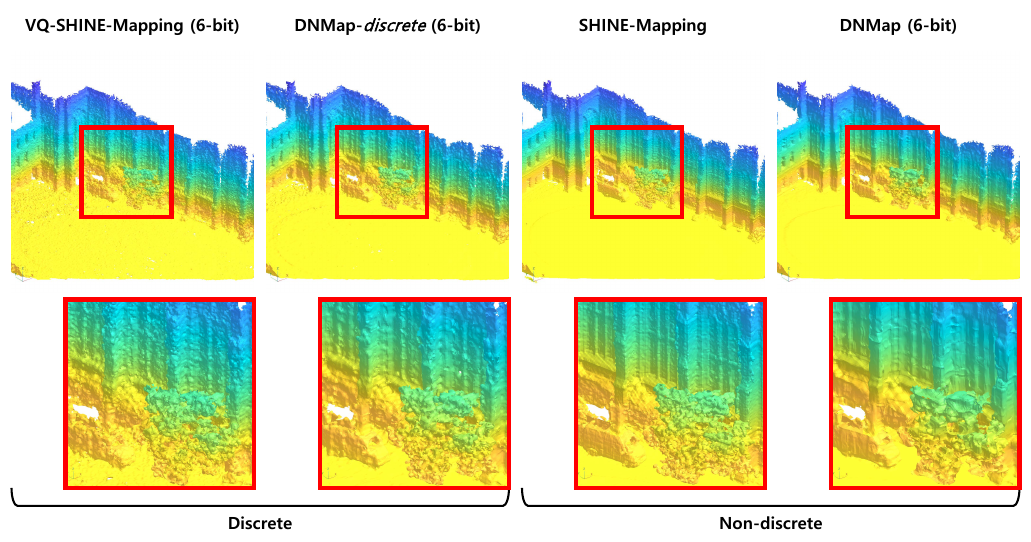}
    \caption{Newer College}
  \end{subfigure}
\caption{Qualitative comparisons on (a) MaiCity~\cite{vizzo2021icra} and (b) Newer College~\cite{ramezani2020newer} datasets. The highlighted areas of the figures are shown below.}
\label{fig:qual}
\end{figure*}

Qualitative comparisons are illustrated in Fig.~\ref{fig:qual}. As shown in the figure, our DNMap reconstructs the shape of the structure as well as SHINE-Mapping. This implies that the proposed decomposition-based representation can capture repetitive and representative structural cues in a large-scale 3D environment, and each discrete embedding is located in an optimal region. Furthermore, our DNMap reconstructs a smoother mesh than SHINE-Mapping, as shown in the highlighted areas. We can confirm that our proposed discrete representation is more robust to noise than the continuous representation.

\subsection{Incremental Mapping}
\label{sec:incre}

\begin{figure}[ht]
\centering
    \centering
    \includegraphics[width=.9\linewidth]{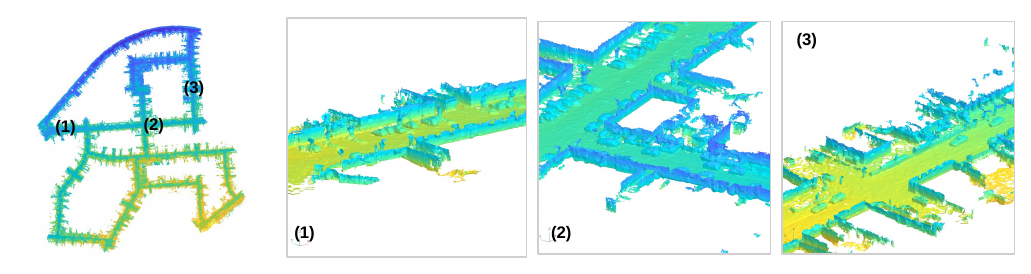}
\caption{The reconstruction result of incremental mapping on KITTI~\cite{kitti} odometry dataset sequence 00 using the component vector set and the decoder pre-trained at Newer College dataset~\cite{ramezani2020newer}.}
\label{fig:incre}
\end{figure}

The reconstruction of incremental mapping is illustrated in Fig.~\ref{fig:incre}. For incremental mapping we adopt replay-based approach \cite{Sucar_2021_ICCV, ortiz2022isdf, Yan_2021_ICCV} that stores the previous data pairs and trains them with current observed data pairs. We perform mapping on the KITTI~\cite{kitti} odometry dataset 00 sequence using the component vector set and the decoder pre-trained at Newer College dataset. As shown in the figure, DNSMap is verified to capture repetitive and representative patterns of the structure well without additional supervision. 

\begin{table}[t]
\caption{Ablation Study. The F-score is reported for every 10000 iterations. EI denotes efficient implementation.}
\label{tab:ablations}
\scriptsize
    \begin{center}
        \begin{tabular}{lcccccccccr}
                \toprule                
                   &\quad\quad\quad\quad &\quad\quad & \multicolumn{5}{c}{F-Score $\uparrow$} &\quad\quad\quad\quad & \multicolumn{2}{c}{it/sec $\uparrow$}  \\ \cmidrule{3-11} 
                 & &    & 10k& 20k& 30k& 40k& 50k && \textit{basic} & \multicolumn{1}{c}{EI}     \\ \midrule
                SHINE-Mapping & & &91.8& 91.9& 91.9& 91.8 & 91.8 &  & 30.7 & \multicolumn{1}{c}{-} \\ 
                DNMap (8-bit) & & &91.7& 92.1& 92.1& 92.2& 92.2 &                         & 23.9 & 27.1(1.13$\times$) \\ \midrule
                $B=4$\\
                \textit{Indexing} & &  &88.1& 89.1& 89.2& 89.3& 88.9&                     & 25.3 & 28.2(1.11$\times$) \\
                \textit{Indexing-Gumbel} & &  &86.4& 86.5& 86.6& 87.2& 86.9 &     & 24.3 & 27.3(1.12$\times$) \\
                \textit{Decomposition-naive} & & &86.7& 87.5& 88.2& 88.2& 88.2 &   & 27.1 & 32.5(1.20$\times$) \\
                \textit{Decomposition} & &  & 87.6& 88.6& 88.9& 88.9& 89.0 &                &25.8 & 30.5(1.18$\times$) \\
                \midrule
                $B=6$\\
                \textit{Indexing} & &  &89.1& 89.7& 89.7& 89.7& 89.7&                     & 20.6 & 22.1(1.07$\times$) \\
                \textit{Indexing-Gumbel} & &  &74.0& 80.4& 83.1& 84.3& 81.6 &     & 18.5 & 19.9(1.08$\times$) \\
                \textit{Decomposition-naive} & & &87.3& 88.5& 88.7& 88.4& 88.7 &   & 26.6 & 31.5(1.18$\times$) \\
                \textit{Decomposition} & &  & 88.7& 89.8& 90.0& 89.9& 89.8 &                &25.9 & 29.9(1.15$\times$) \\ 
                \midrule
                $B=8$\\                
                \textit{Indexing} & &  &89.0& 89.4& 89.5& 89.5& 89.6&                     & 12.3 & 12.4(1.01$\times$) \\
                \textit{Indexing-Gumbel} & &  &77.8& 83.0& 80.5& 83.5& 85.0&      & \multicolumn{1}{r}{9.8}  & 9.9(1.01$\times$) \\
                \textit{Decomposition-naive} & & &88.7& 89.7& 90.0& 89.8& 90.0 &   & 26.3 & 30.4(1.16$\times$) \\
                \textit{Decomposition} & &  & 88.7& 90.2& 90.3& 90.3& 90.3 &                &25.6 & 29.5(1.15$\times$) \\
             
                \bottomrule
		\end{tabular}
	\end{center}
\end{table}

\subsection{Ablation Studies (Tab.~\ref{tab:ablations})}
We conduct ablation studies on the proposed method to analyze our proposed method by comparing \textit{Indexing}, \textit{Indexing-Gumbel}, \textit{Decomposition-naive}, and \textit{Decomposition}. \textit{Indexing} indicates VQ-SHINE-Mapping that learns indexing using Softmax function \cite{zhong2023icra,10.1145/3528233.3530727}. \textit{Indexing-Gumbel} is a method of providing randomness to indexing in VQ-SHINE-Mapping using the Gumbel-Softmax function \cite{jang2016categorical,zhong2023icra,10.1145/3528233.3530727}. \textit{Decomposition-naive} is a naive version of our decomposition-based discrete representation where the composition process is Eq.~(\ref{eq:naive_composition}). \textit{Decomposition} indicates our decomposition-based discrete representation where the composition process is Eq.~(\ref{eq:composition}).

All ablation studies are trained with some variation on the experimental setting in Section~\ref{sec:imp_detail}: $N_F=4$, $N_S=8$, and $s=0.05$ for data preparation; 50000 iterations with 4096 batch size for training. We report the F-score for every 10000 iterations and the training speed (it/sec).

\subsubsection{Analysis on Neural Discrete Representation Learning.}
From the table, \textit{Indexing} shows degraded performance as the bitwidth increases, and the learning speed decreases significantly. In our experience, indexing-based representation fails to utilize all embeddings when the size of the embedding space becomes too large and falls into a local minimum. Moreover, the performance of \textit{Indexing-Gumbel} is very poor, likely because the introduced randomness confuses the feature volume and the decoder that optimizes neural fields. \textit{Decomposition-naive} has the fastest learning speed but relatively low performance, whereas Sigmoid-\textit{proposed}, which can express a more diverse embedding space, outperforms all other methods.

\subsubsection{Analysis on Training Speed.}
From the table, we can see that decomposition-based representation can optimize the discrete embedding space with faster training speeds compared to other methods. Additionally, the efficient implementation (EI) described in Section~\ref{sec:efficient_implementation} significantly boosts training speed. However, if accessing the parameter set representing the embedding space is not the main cause of slowing down the learning speed, there is no significant effect.

\section{Discussion}
Although our proposed decomposition-based discrete representation shows improved performance and storage efficiency compared to existing representations, it still has weaknesses. While our DNMap is robust to noise and specialized for learning structural cues in large-scale 3D environments, it still has difficulties representing thin surfaces, which are generally a weak point in implicit neural representations. To solve this problem, we can reduce the voxel size to represent more details and increase the height of the octree, but this is not a fundamental solution. Therefore, our future work will focus on maintaining local structural details while still being memory efficient in large-scale environments.

\section{Conclusion}
In this paper, we present a novel 3D mapping framework named Decomposition-based Neural Mapping (DNMap) to capture repetitive and representative patterns in a large-scale environment in a storage-efficient manner. To this end, we adopt an implicit neural discrete representation based on a decomposition strategy that learns the decomposition and composition of the discrete embedding space. First, we decompose each discrete embedding into component vectors that are shared across the embedding space. Then, we learn the composition indicators and store them in an octree-based structure. All parameters are simultaneously optimized without additional supervision. With these techniques, we significantly reduce the storage of local features in large-scale outdoor datasets while preserving mapping accuracy.

\subsubsection*{Acknowledgments.}
This work was supported by Korea Evaluation Institute Of Industrial Technology (KEIT) grant funded by the Korea government(MOTIE) (No.20023455, Development of Cooperate Mapping, Environment Recognition and Autonomous Driving Technology for Multi Mobile Robots Operating in Large-scale Indoor Workspace)

%
%
\bibliographystyle{splncs04}
\bibliography{main}
\end{document}